\documentclass[a4paper,french]{rnti}

\usepackage{graphicx}
\usepackage[T1]{fontenc}
\usepackage[utf8]{inputenc}

\usepackage{bigfoot}

\usepackage{array}
\newcolumntype{L}[1]{>{\raggedright\let\newline\\\arraybackslash\hspace{0pt}}m{#1}}
\newcolumntype{C}[1]{>{\centering\let\newline\\\arraybackslash\hspace{0pt}}m{#1}}
\newcolumntype{R}[1]{>{\raggedleft\let\newline\\\arraybackslash\hspace{0pt}}m{#1}}

\makeatletter
\newcommand\footnoteref[1]{\protected@xdef\@thefnmark{\ref{#1}}\@footnotemark}
\makeatother

\newcounter{daggerfootnote}
\newcommand*{\daggerfootnote}[1]{%
    \setcounter{daggerfootnote}{\value{footnote}}%
    \renewcommand*{\thefootnote}{\fnsymbol{footnote}}%
    \footnote[2]{#1}%
    \setcounter{footnote}{\value{daggerfootnote}}%
    \renewcommand*{\thefootnote}{\arabic{footnote}}%
    }

\usepackage{url}

\titrecourt{Emotional Speech Recognition
with Pre-trained Deep Visual Models}

\nomcourt{W. Ragheb et al.}

\titre{Emotional Speech Recognition with Pre-trained Deep Visual Models}

\auteur{Waleed Ragheb 	\daggerfootnote{\label{EQU}Authors with equal contribution to this work} \affil{1}\affilsep\affil{2},
        Mehdi Mirzapour  \footnoteref{EQU}	\affil{3}\\
        Ali Delfardi\affil{4},
        Hélène Jacquenet\affil{3},
         Lawrence Carbon\affil{3}}

\affiliation{
    \affil{1} LIRMM, Univ Montpellier, CNRS, Montpellier, France\\
          waleed.ragheb@lirmm.fr\\ 
   \affil{2} IT Dept,
    Faculty of Computers and Artificial Intelligence,\\ Cairo University, Egypt\\

    \affil{3}ContentSide, R\&D Dept, Lyon, France\\
          \{first.last\}@contentside.com\\ 
          
    \affil{4} CS Dept, Isfahan University, Isfahan, Iran\\
         alidelfardi@eng.ui.ac.ir\\

 }

\resume{%
In this paper, we propose a new methodology for emotional speech recognition using visual deep neural network models. We employ the transfer learning capabilities of the pre-trained computer vision deep models to have a mandate for the emotion recognition in speech task. In order to achieve that, we propose to use a composite set of acoustic features and a procedure to convert them into images. Besides, we present a training paradigm for these models taking into consideration the different characteristics between acoustic-based images and regular ones. In our experiments, we use the pre-trained VGG-16 model and test the overall methodology on the Berlin EMO-DB dataset for speaker-independent emotion recognition. We evaluate the proposed model on the full list of the seven emotions and the results set a new state-of-the-art. 
}

\begin{document}

%
\section{Introduction}
With the information technology revolution, it became mandatory -- not just an option -- for many computer systems to express and recognize effects and emotions to attain creative and intelligent behavior. The main purpose is to understand the emotional states expressed by the human subjects so that personalized responses can be delivered accordingly. Humans are still way ahead of machines in detecting and recognizing the different types of effects including emotions \citep{alu2017voice}. Therefore, Emotional Intelligence (EI) is deemed as the turning point from moving from the narrow definition of Artificial  Intelligence (AI) to a more general humanized AI. Speech signals are considered one of the main channels in human communications. Naturally, humans could effectively recognize the emotional aspects of speech signals. The emotional state of the speech will not change the linguistics of the uttered speech, but it reflects many of the speaker's intents and other latent information about the mental and physical state and attitude \citep{narayanan2013behavioral}. Therefore, empowering computer systems with speech emotional recognition features can have a significant impact on personalizing the user experience in many applications and sectors such as  marketing, healthcare, customer satisfaction, gaming experience improvement, social media analysis and stress monitoring. \citep{nassif2019speech, proksch2019testing, rouhi2019emotify}. 

Earlier emotional speech recognition had some processes in common with automatic speech recognition. It involved many feature engineering steps that may play a substantial role in model selection and training paradigm. Acoustical speech features reported in the literature could be categorized into continuous, qualitative, spectral, and temporal features \citep{bandela2017stressed}. At the time, most of the models were classical machine learning and statistical models. Most of these models train from scratch on a varied set of features or the original speech signal itself. Different pre-trained models have been released and become substantially available for many applications in Computer Vision and Natural Language Processing. 

As for emotional speech recognition, some pre-trained transferable models such as speechVGG \citep{beckmann2019speech} have been introduced which act as feature extractor for different speech processing tasks. Although speechVGG has got its inspiration from VGG \citep{simonyan2014very}-- a well-known computer vision architecture-- it is trained from scratch with the LibriSpeech dataset \citep{librispeech}. We mainly focus on how an existing pre-trained computer vision model, such as VGG, can efficiently be fine-tuned in a different domain such as emotional speech recognition. This can reduce the cost of further expensive and exhaustive training for new domains and be beneficial for practical and industrial use cases.

In this work, we present an experimental study using one of the most powerful pre-trained visual models VGG to tackle the aforementioned problem. Our proposed methodology is: (i) to present a novel order of frequency-domain voice features that transform the speech acoustic signals into compound ready-to-use 3D images for existing pre-trained computer vision models; (ii) to apply simple signal-level and frequency domain voice-level data augmentation techniques;  (iii) to introduce simple, and yet efficient mini-batch padding technique; and finally, (iv) to fine-tune the VGG-16 (with batch-normalization) pre-trained model on classical image classification tasks. We have applied the proposed configurations and some of their variants on one of the most well-known datasets for emotional recognition (Berlin EmoDB \citep{burkhardt2005database}) and the results are very competitive to the state-of-the-art and outperform many strong baselines. Our implementation is made available for public\footnote{https://github.com/mehdi-mirzapour/Emotional\_Speech\_Recognition}.

The paper is organized as follows: in section \ref{RW} we present a literature review and the related works. In section \ref{PM}, we introduce the proposed methodology including the considered acoustic features and all the variants of the used models. Section \ref{Exp} addresses all the experimental setups and the results followed by brief discussions. Finally, we conclude the study and experiments in section \ref{Conc}.

\section{Related Works}\label{RW}
Traditional models proposed for emotional speech recognition and classification are based on the same models used for automatic speech recognition like HMM, GP, SVM ... etc \citep{svmHass, lin2005speech, azmy2013arabic}. These models involve extensive feature engineering steps that are sensitive and may significantly affect the structure of the entire method \citep{pandey2019deep}. With the development of the deep learning models, the speech recognition systems benefited from the end-to-end learning paradigm. This enables the model to learn all the steps from the input to the final output simultaneously including feature extraction. Similarly, the emotional models have followed the same course. There is a lot of effort and research on employing these algorithms to recognize emotions from speech. More specifically, some of these models used the ability of Convolutional Neural Networks (CNN) to learn features from input signals  \citep{bertero2017first, mekruksavanich2020negative}. Another type of model makes use of the sequential nature of the speech signals and utilized Recurrent Neural Networks (RNN) architectures like long short-term memory (LSTM) \citep{tzinis2017segment, fayek2017evaluating}. Some models combined both types of architectures like in ConvLSTM \citep{kurpukdee2017speech}.

\begin{figure}
   \centering
  \includegraphics[width=1.0\columnwidth]{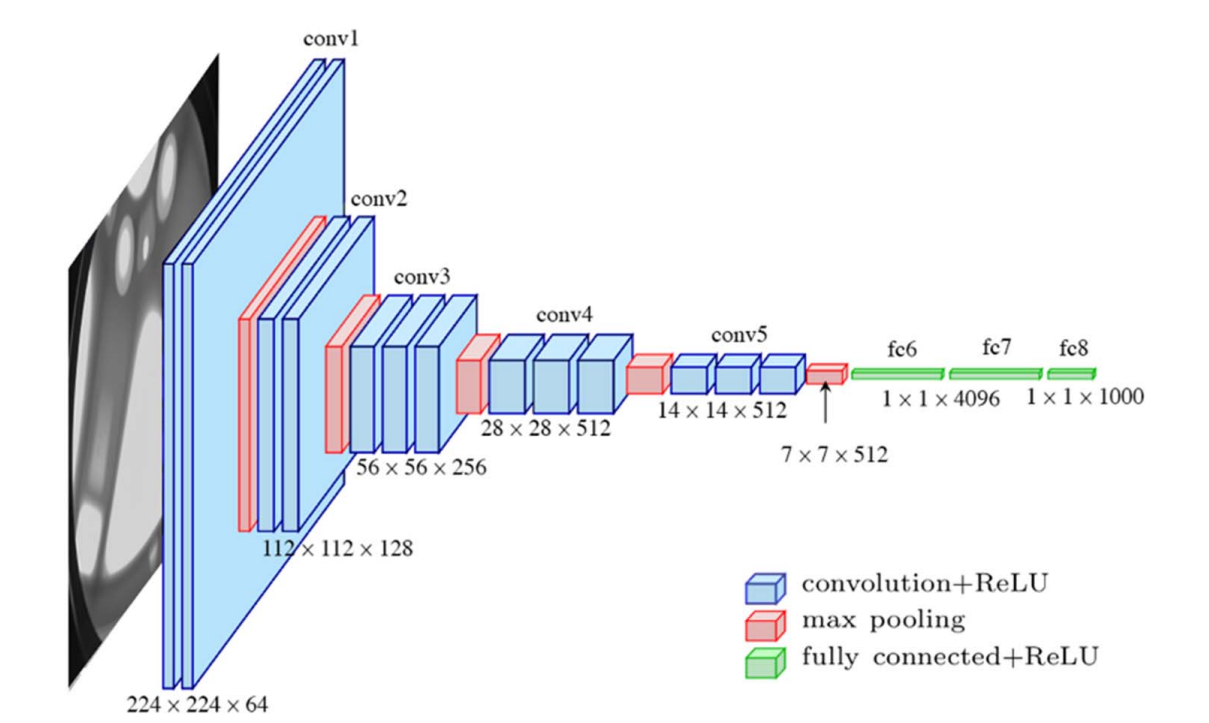}
  \caption{The original VGG-16 architecture \citep{simonyan2014very}}
  \label{fig_vgg}
\end{figure}  

Recently, there has been a breakthrough improvement in the transfer learning capabilities of deep models with the powerful pre-trained visual models like AlexNet, VGG, Yolo-2 ... etc \citep{cvPre}. The main idea is to train these models for large and different image classification tasks and transfer the feature selection parts of these models to be used in the downstream tasks. Figure \ref{fig_vgg} shows an example of these models and the one used in our experiments (VGG). This complies with the fact that speech signals could be represented as visual features. For instance, the  Mel frequency Cepstral coefficient (MFCC), Log Frequency Power Coefficients (LFPC), and Log Mel Spectrogram could be considered as 2D/3D images that could carry emotion-related information \citep{wang2014feature}. This will permit us to take advantage of the pre-trained visual models to extract visual features presented in the input acoustic features without the need for large datasets in an indirect supervision fashion. The work presented in this paper is to some extent related to a previous work in \citep{zhang2017speech}. The authors use only Log Mel Spectrogram on three channels of deltas as the input features and a pre-trained AlexNet \citep{krizhevsky2012imagenet} as the visual model. This model extracts the visual feature representation of the input and then involves a linear SVM model for the target classification task.

\section{Proposed Methodology}\label{PM}

In this section, we present the preprocessing steps and the proposed set of acoustic features that we used. Besides, we introduce the architecture of the visual model applied in the experiments with all the considered variants. 

\subsection{Acoustic Features}

We tried different types of acoustic features to get the proper representation of the speech signals in the form of images. The frequency-domain features reveal a promising behavior. More specifically, we used the same spectral features reported in \citep{issa2020speech} and utilized the Librosa library \citep{mcfee2015librosa} for the features extraction process. Additionally, we added one more feature and proposed a new method for integrating all these features into an image representation of the input speech signal.  The complete set of used features are:
\begin{enumerate}

\item Mel-frequency cepstral coefficients (MFCCs)
\item Mel-scaled spectrogram.
\item Power spectrogram or Chromagram 
\item Sub-bands spectral contrast 
\item Tonal centroid features (Tonnetz) 
\item Average of the mel-scaled spectrogram of the harmonic and percussive components\\
 \end{enumerate}
 
After removing small silence (pauses) in the speech signals, we compose the images by a normalized concatenation of all the previous features in three different channels like the Red-Green-Blue (RGB) decomposition of an image. In contrast to \citep{issa2020speech}, we did not aggregate the time scale features for achieving a fixed sized vector. This is a crucial decision for fine-tuning the VGG models since they are data-greedy and aggregating the time-scale information (by summing or averaging functions) will eliminate a considerable amount of useful patterns resulting in accuracy reduction of our model. Nevertheless, this strategy has a side effect for some architectures: the resulting 3D images will vary in time-dimension axis. Our case study on EMO-DB dataset shows the size (3, X, 230) in which X varies in time axis between 50 and 700 depending on audio signal sizes. This, in principle, will not affect our model since VGG models require a minimum input size of the 3x32x32 which fits well with our settings. Figures (\ref{example_aug}a)-(\ref{example_aug}b) show two examples of original speech visual features happiness and anger, respectively.

It is worth mentioning that the order of the features has an important role in getting an acceptable accuracy. To find the optimum order, we experimented with transforming our features to vectors by averaging values in the time axis, and then we have fed all the permutations of these compact vector features to a very simple logistic regression model classifier.  We finally selected only a few candidates with the same accuracy. To make the final decision on the orders of features, we fine-tuned our VGG-16 model according to the shortlisted orders of features to get the best performance. This has given us the practical ground to find the best permutations of the order of the features.

\subsection{Model Training}

For the visual deep model, we used the pre-trained version of 16-layer VGG architecture (VGG16) \citep{simonyan2014very} which is proven to get the best performance in ILSVRC-2014\footnote{ImageNet Large Scale Visual Recognition Challenge 2014 - https://www.image-net.org/challenges/LSVRC/2014/} and became very popular in many image classification problems. The feature extraction layer groups are initialized from the pre-trained model weights, however, the classification layer group is fine-tuned from scratch with random weights. Moreover, we used the batch normalization variants of the VGG model for its effective regularization effect especially with training batches of smaller sizes. When creating the batches and before extracting the acoustic features, we applied signal-level padding to the maximum signal length in each training batch.

\begin{figure}
   \centering
  \includegraphics[width=120mm,scale=0.7]{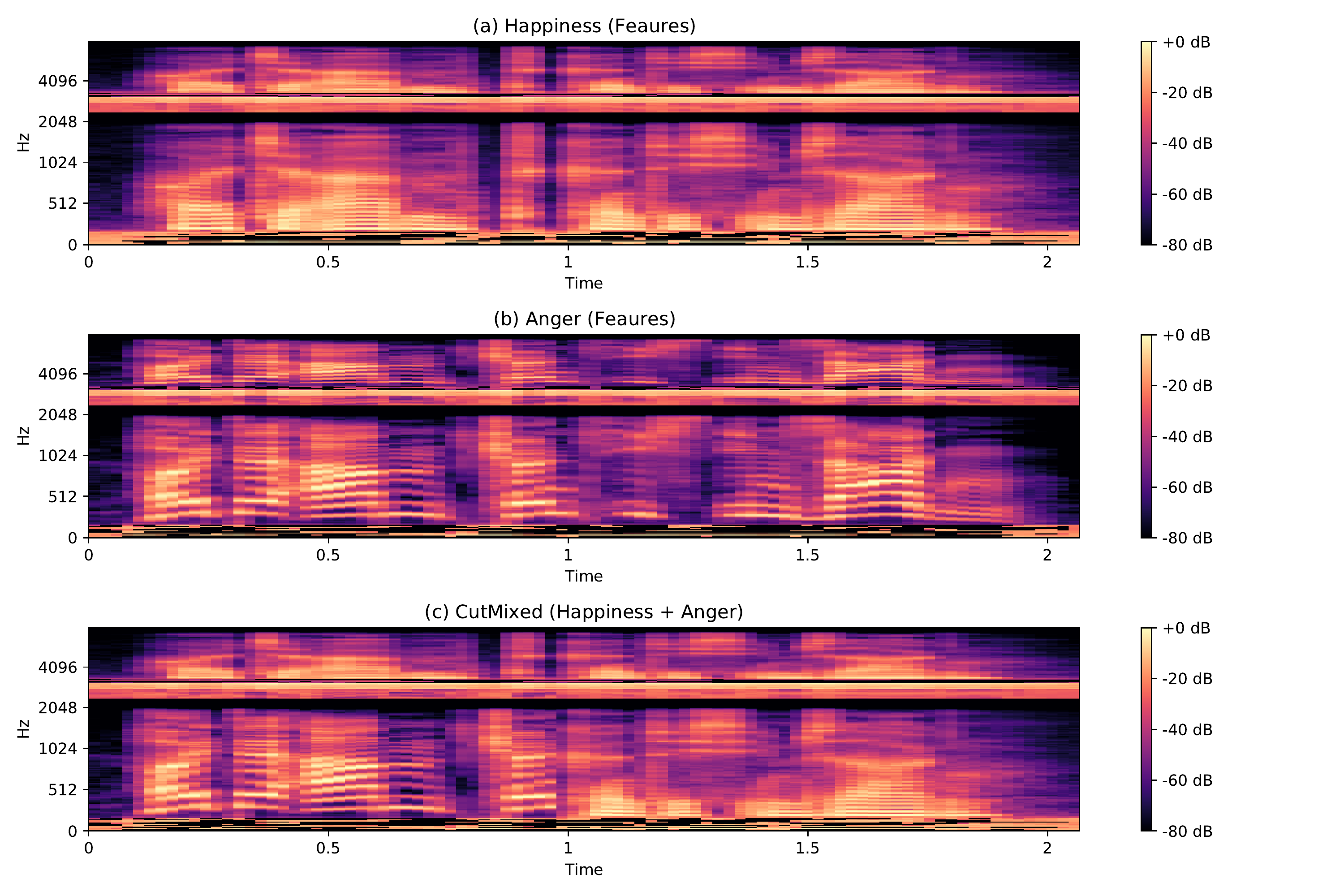}
  \caption{Two examples of original speech visual features (happiness and anger) representation before and after \textit{"CutMix"}}
  \label{example_aug}
\end{figure}

Most of the available emotional speech datasets are small-sized. Accordingly, this was the primary motivation for relying on pre-trained models. In addition to that, and as most deep visual model training, we applied data augmentation. The idea here is not as simple as regular transformation done with the images - rotation, translation, flipping, cropping ... etc. Our proposed images in the considered problem are special types of images, so standard visual augmentation techniques will not be useful. Hence, we applied \textit{"CutMix"}; \citep{yun2019cutmix} a special augmentation and regularization strategy in which patches are cut and pasted among training images where the ground truth labels are also mixed proportionally in the area of the patches. \textit{"CutMix"} efficiently uses training pixels and retains the regularization effect of regional dropout. Figure \ref{example_aug} shows two examples of original speech visual features (happiness and anger) representation before and after \textit{"CutMix"}.

As a regular classification deep learning model, we used cross entropy loss function with Adam optimizer \citep{kingma2014adam}. We employed a learning rate of $10^{-5}$ and a batch size of 16.

\section{Experiments}\label{Exp}

In this section, we first present the used datasets. Then, we show the results of the proposed methodology before we end with some discussions.
\subsection{Dataset}
\begin{figure}
   \centering
  \includegraphics[width=110mm,scale=0.7]{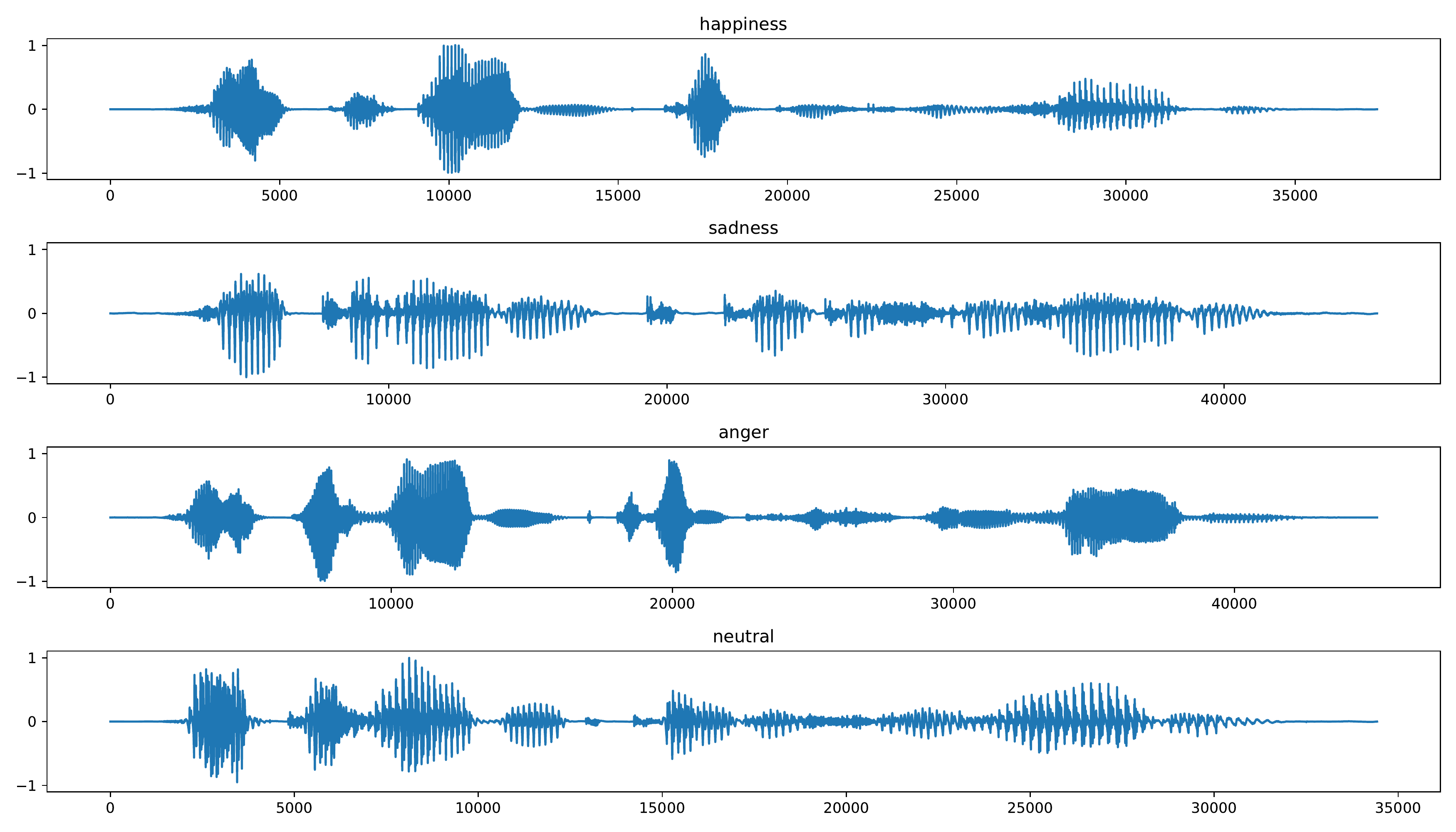}
  \caption{Example of a same speech uttered in different emotional states by a same speaker}
  \label{exp_speech}
\end{figure}

In these experiments, we used Berlin EMO-DB \citep{burkhardt2005database} with its speaker-independent configuration that contains a total of 535 emotional speech German utterances. Each speech signal is classified into one of the seven classes of emotions (Fear, Sadness, Disgust, Anger, Boredom, Neutral, and Happiness). The dataset contains some examples of the same text spoken with different emotions (see figure \ref{exp_speech}). The dataset contains speech signals corresponding to different speakers.

We split the dataset randomly into training and testing sets and preserve almost the same distributions of different emotions across both datasets as summarized in table~\ref{ds_stat}.

\begin{table}[htp]
\centering 

\begin{tabular}{C{3cm} R{2cm} R{2cm}}
\hline\hline 

 Classes & Train & 	Test \\  
\hline 
Fear & 55 &	14  \\
Sadness & 50 &	12   \\ 
Disgust & 37 &	9  \\
Anger & 102 &	25 	\\ 
Boredom    & 65 &	16  \\
Neutral  & 64 &	15   \\ 
Happiness & 56 &	15   \\ 

\hline 
Total & 429 & 106 \\
\hline
\end{tabular}
\caption{Summary of the Berlin EmoDB dataset}
\label{ds_stat}
\end{table}

\subsection{Results}
We tested the overall methodology described in section \ref{PM} to measure the classification performance on the test dataset. We use the accuracy metrics to enable the head-to-head comparison with the SOTA models and other reported baselines results. Moreover, we tested 6 possible variants of the model to enhance the ablation analysis of the proposed methodology components. The proposed variants and their definition are described as:

\begin{enumerate}
\item Model-A: The complete model as described in section  \ref{PM}
\item Model-B: The model without batch normalization
\item Model-C: Excluding \textit{CutMix} augmentation strategy 
\item Model-D: Excluding signal-level augmentation
\item Model-E: Excluding both signal-level and \textit{CutMix} augmentations
\item Model-F: Applying the model excluding mini-batch padding
\end{enumerate}

Table \ref{res_var} presents the results of all these variants which validate that the best performing model is corresponding to the complete proposed methodology. Besides, we compare this model (Model-A) to a set of strong baseline including the SOTA best-reported results for the considered dataset. We show this comparison in table \ref{res_sota}. Furthermore, we present the confusion matrix of our best model concerning all different classes in figure \ref{fig_CM}.

\begin{table}[htp]
\centering 

\begin{tabular}{C{3cm} R{2cm}}
\hline\hline 
 Model & Accuracy (\%) \\  
\hline 
Model-A & \textbf{87.73}  \\
Model-B & 83.96 \\ 
Model-C & 83.02 \\
Model-D & 81.13 \\ 
Model-E & 76.42 \\
Model-F & 69.81 \\
\hline
\end{tabular}
\caption{Classification accuracy of the proposed model variants}
\label{res_var}
\end{table}

\subsection{Discussions}

\begin{figure}
   \centering
  \includegraphics[width=100mm,scale=0.8]{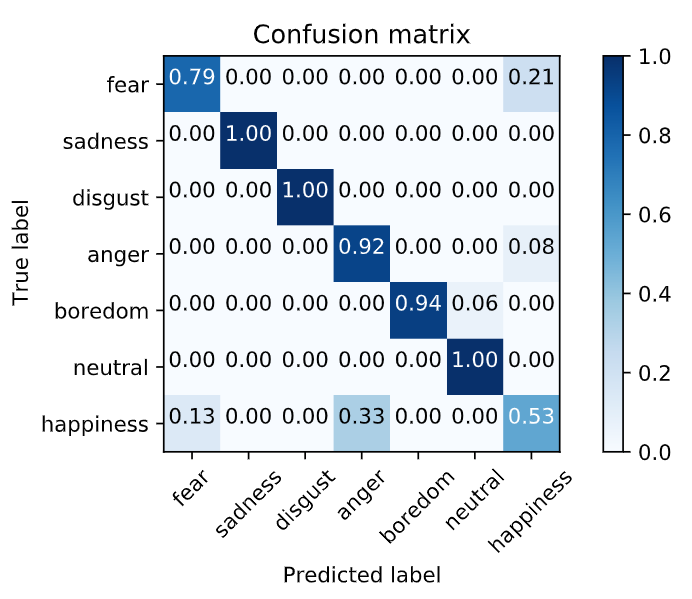}
  \caption{Confusion matrix of Model-A}
  \label{fig_CM}
\end{figure}

By applying the overall methodology described in section \ref{PM}, we have reached our best model (Model-A) with the 87.73 accuracy measure. Moreover, six other variants of the model (as in table \ref{res_var}) have been taken into consideration to enhance the ablation analysis of the proposed methodology. A quick analysis shows that mini-batch padding plays the most significant role as Model-F indicates around -18 percent drop in the accuracy by excluding it. Excluding both signal-level and \textit{CutMix} augmentations (Model-E) can reduce the performance of our model by around 11 percent. This indicates the role of two-sided augmentations of the model in the signal and image levels. Analysis of Model-C and Model-D show that stand-alone signal-level augmentation is just a slightly better component than stand-alone \textit{CutMix} augmentation as the signal-level augmentation is only 2 percent ahead of \textit{CutMix} augmentation in terms of accuracy measure. As discussed earlier, both augmentations can consistently strengthen each other. Model-B shows the importance of using batch-normalization to make the final model better to some extent (batch-normalization was later introduced and added to VGG architectures). Table \ref{res_sota} shows our model outperformed the previous state-of-the-art results and many strong base-lines. Figure \ref{fig_CM} shows the detailed confusion matrix for our best model result. It is worth mentioning that our analysis shows the importance of applying different components altogether (and not just only fine-tuning VGG-16) to outperform the state-of-the-art results. 

\begin{table}[htp]
\centering 

\begin{tabular}{C{9cm} R{2cm}}
\hline\hline 
 Model & Accuracy (\%)  \\  
\hline 
Badshah et. al. \citep{7883728} & 52.00  \\
Wang et. al. \citep{wang2015speech} & 73.30  \\ 
Lampropoulos et. at. \citep{6274410} & 83.93 \\
Huangb et. al. \citep{10.1145/2647868.2654984} & 85.20 \\ 
Wu et. al. \citep{WU2011768} & 85.80\\
Issa et. al. \citep{issa2020speech} $^*$ & 86.20\\
\hline
Our best model (Model-A) & \textbf{87.73}\\
\hline
\end{tabular}
\caption{Classification accuracy of our best performing model compared to other model including the SOTA (*) model }
\label{res_sota}
\end{table}

\section{Conclusions}\label{Conc}
Speech is one of the most preferred means in human communications. With the recent advances in speech technology and human/machine interaction, emotional speech recognition systems play an important role in bringing out emotionally intelligent behavior tasks. This study has focused on the VGG-16 (with batch normalization) pre-trained computer vision model and we have highlighted efficient components for fine-tuning VGG-16 for emotional speech recognition. This has been achieved by applying a novel order of frequency-domain voice features represented as ready-to-use 3D images; signal-level and frequency-domain voice-level data augmentation techniques; and finally simple, and yet efficient, mini-batch padding technique. We have outperformed the previous state-of-the-art results and many strong baselines.

The work presented in this paper could be extended to include more pre-trained computer vision deep models such as ResNet \citep{he2016deep}, EfficientNet\citep{tan2019efficientnet}, ViT\citep{dosovitskiy2020image} and Inceptionv3 (GoogLeNet) \citep{szegedy2016rethinking}. Besides, extensive experiments can be performed on other emotional datasets like LSSED \citep{fan2021lssed}, IEMOCAP \citep{busso2008iemocap}, and RAVDESS \citep{livingstone2018ryerson} . Moreover, it could be interesting to include other modalities for emotional recognition like text, images and videos.

\bibliographystyle{rnti}
\bibliography{DL4NLP_paper}

\end{document}